%% file: main.tex
\documentclass[10pt,twocolumn,letterpaper]{article}

\usepackage[pagenumbers]{cvpr}  

\input{preamble}

\usepackage[pagebackref,breaklinks,colorlinks,linkcolor=blue,citecolor=blue,urlcolor=blue]{hyperref}
\usepackage{multirow}
\usepackage[table,xcdraw]{xcolor}
\usepackage{colortbl}
\usepackage{threeparttable}

\title{Topology-Agnostic Animal Motion Generation from Text Prompt}

\author{
    Keyi Chen\textsuperscript{1,*} \quad
    Mingze Sun\textsuperscript{1,*} \quad 
    Zhenyu Liu\textsuperscript{1} \quad
    Zhangquan Chen\textsuperscript{1} \quad
    Ruqi Huang\textsuperscript{1,\dag}
    \\[1em]
    \textsuperscript{1}Tsinghua Shenzhen International Graduate School, China
    \\[0.5em]
    {\tt\small \{chenky24,sunmz21, liuzheny25,czq23\}@mails.tsinghua.edu.cn, ruqihuang@sz.tsinghua.edu.cn}
    \\[0.2em]
    {\small \textsuperscript{*}Equal contribution \quad \textsuperscript{\dag}Corresponding authors}
}

\begin{document}

\twocolumn[{

\renewcommand\twocolumn[1][]{#1}
\maketitle


\begin{center}
    \captionsetup{type=figure}
    \centerline{\includegraphics[width=\linewidth]{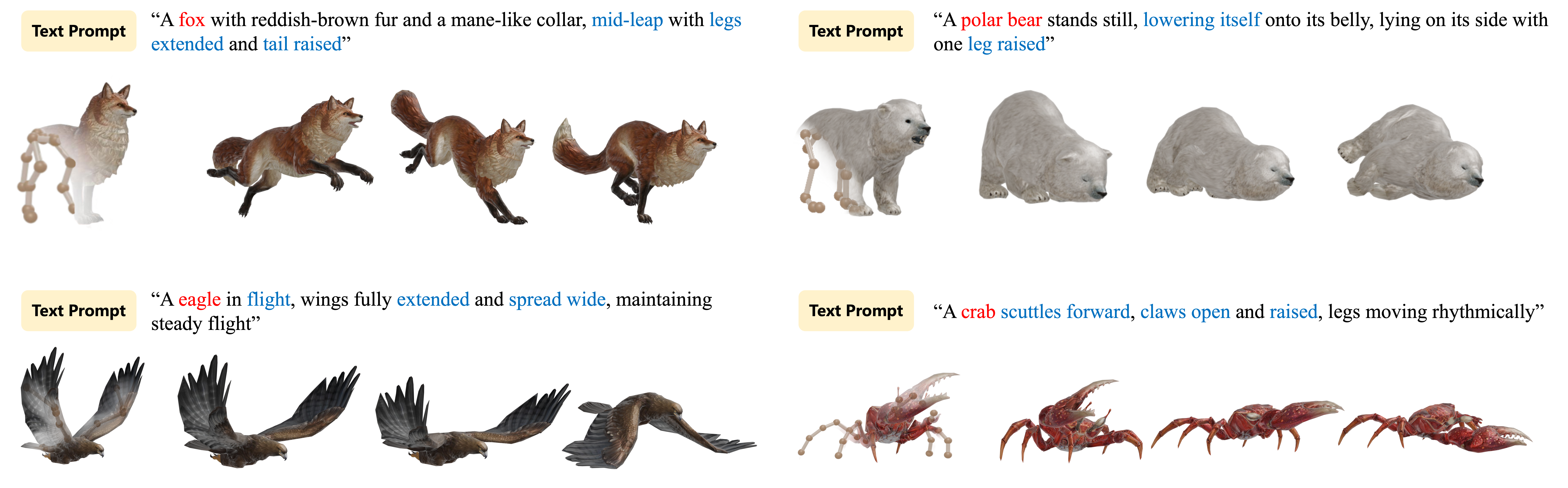}}
    \caption{
    Given an arbitrary skeletal topology and a corresponding text prompt, our method drives the skeleton to produce realistic, high-quality motions that align with the textual description.
    }
    \label{fig:teaser}
\end{center}
}]

\input{sec/0_abstract}    
\input{sec/1_intro}

\input{sec/2_related}

\input{sec/3_dataset}

\input{sec/4_method}
\input{sec/5_experiments}
\input{sec/6_conclusion}
{
    \small
    \bibliographystyle{ieeenat_fullname}
    \bibliography{main}
}


\end{document}

%% file: preamble.tex
\usepackage{threeparttable}



%% file: sec/0_abstract.tex
\begin{abstract}

Motion generation is fundamental to computer animation and widely used across entertainment, robotics, and virtual environments. 
While recent methods achieve impressive results, most rely on fixed skeletal templates, which prevent them from generalizing to skeletons with different or perturbed topologies. 
We address the core limitation of current motion generation methods—the combined lack of large-scale heterogeneous animal motion data and unified generative frameworks capable of jointly modeling arbitrary skeletal topologies and textual conditions.
To this end, we introduce OmniZoo, a large-scale animal motion dataset spanning 140 species and 32,979 sequences, enriched with multimodal annotations.
Building on OmniZoo, we propose a generalized autoregressive motion generation framework capable of producing text-driven motions for arbitrary skeletal topologies. 
Central to our model is a Topology-aware Skeleton Embedding Module that encodes geometric and structural properties of any skeleton into a shared token space, enabling seamless fusion with textual semantics. 
Given a text prompt and a target skeleton, our method generates temporally coherent, physically plausible, and semantically aligned motions, and further enables cross-species motion style transfer.

\end{abstract}

%% file: sec/1_intro.tex
 \section{Introduction}
\label{sec:intro}

Motion generation lies at the core of modern computer animation and plays a pivotal role in a wide range of downstream applications, including digital entertainment, robotics, and virtual reality. 
Realistic motion is the foundation of lifelike character behaviors and expressive storytelling. 
Traditionally, high-quality motions are crafted manually by professional animators through tedious keyframe editing and motion capture retargeting. 
However, such hand-crafted pipelines are time-consuming and labor-intensive, creating a pressing need for automatic motion generation methods that can efficiently produce diverse, controllable motions with minimal human intervention.

In recent years, learning-based motion generation has attracted growing research attention, enabling the automatic synthesis of human and animal motions directly from data. 
Most existing approaches, however, are designed under the assumption of a fixed template. 
Representative examples include SMPL~\cite{loper2023smpl} for human motions~\cite{guo2024momask, sun2025beyond, xu2025combo, lu2025scamo} and other manually-defined templates~\cite{zuffi20173d, wang2025animo} for animal motion systems~\cite{yang2024omnimotiongpt, wang2025animo}, which rely on canonical templates to define the skeletal structure. 
While effective for their target species, these template-based methods are inherently limited: even slight topological perturbations—such as missing joints, merged limbs, or additional bones—can break compatibility, making them incapable of generalizing to new skeletons or unseen species. 
This limits the downstream applicability of template-based motion generation methods.

To overcome the rigidity of template-constrained motion models, recent work AnyTop~\cite{gat2025anytop} has made a pioneering attempt to address the problem of motion generation for heterogeneous skeletons. 
By representing motions in a topology-agnostic form, AnyTop demonstrates the feasibility of producing motions for arbitrary skeletons. 
Nevertheless, its unconditional diffusion-based formulation suffers from stochastic generation and limited controllability. 
Moreover, AnyTop lacks semantic conditioning, preventing users from guiding the motion generation process with high-level intent such as natural-language descriptions or desired behavioral styles—capabilities that are essential for scalable and user-friendly animation systems.
To address the issues mentioned earlier, two key challenges must be overcome.
First, there is a lack of large-scale, high-quality animal motion datasets — for example, Truebones Zoo~\cite{studios2022truebones} contains only about $1.4$k motion sequences.
Secondly, current generative frameworks fall short in jointly handling arbitrary skeletal topologies and textual inputs in a unified manner.

In light of these challenges, we take the first step toward semantically controllable motion generation for arbitrary topologies.
We first introduce \textbf{OmniZoo}, a large-scale heterogeneous animal motion dataset covering $140$ species and a total of $32,979$ motion sequences.
Our dataset provides rich multimodal annotations, including textual descriptions ranging from simple to fine-grained motion narratives, rendered videos, skeleton sequences, and corresponding mesh animations, thereby filling an important gap in current motion generation research.

Building upon this foundation, we propose an autoregressive generalized motion generation model.
Given a text prompt and a target skeleton, our model generates temporally coherent and physically plausible motions that are semantically aligned with the textual description and topologically consistent with the input skeleton.
At its core, we design a Topology-aware Skeleton Embedding Module capable of capturing both the spatial geometry and structural topology of arbitrary skeletons, encoding them into a shared token space.
This unified representation allows the model to seamlessly interpret both textual instructions and skeletal structures as conditioning signals, enabling flexible and generalized motion generation.
Extensive experiments across a wide range of species and motion types demonstrate that our framework can generate high-quality, semantically controllable motions for diverse animal species, and further enable cross-species motion style transfer, showcasing strong generalization and versatility across heterogeneous skeletal topologies.

Our contributions can be summarized as follows:
1) We introduce OmniZoo, a large-scale heterogeneous animal motion dataset covering $140$ species and $32,979$ motion sequences with rich multimodal annotations including text, video, skeletons, and meshes.
2) We propose a generalized autoregressive motion generation framework that jointly models text and arbitrary skeletons via a Topology-aware Skeleton Embedding Module, enabling unified and controllable motion generation.
3) Extensive experiments show that our model generates high-quality, semantically controllable motions across diverse species.

%% file: sec/2_related.tex
\section{Related Work}

\subsection{Text-Driven Human Motion Generation}

Text-to-motion generation has recently made significant progress, driven largely by advances in large-scale generative models. 
Existing methods fall into two main paradigms: diffusion-based approaches~\cite{tevet2023human, sun2024lgtm, zhang2024motiondiffuse, ruiz2025mixermdm, zhang2025energymogen}, which synthesize motions via iterative denoising, and autoregressive approaches~\cite{zhang2023generating, pinyoanuntapong2024mmm, zhong2023attt2m, han2025atom, liao2025shape}, which discretize motions into tokens and generate them sequentially in GPT-like fashion.
T2M-GPT~\cite{zhang2023generating} first introduced VQ-VAE–based motion tokenization, while masked modeling frameworks such as MoMask~\cite{guo2024momask} adopt BERT-style token recovery.
Human motion generation has benefited greatly from unified templates like SMPL~\cite{loper2023smpl}, enabling large-scale motion capture and high-quality generation. 
However, such template-based representations have limited expressiveness, making it difficult for existing methods to generalize beyond their predefined skeletal structures.

\subsection{Animal Motion Generation}
Extending text-driven motion generation to animals introduces distinctive challenges. The diversity of animal morphologies and species-specific behavioral patterns fundamentally differs from human motion. OmniMotionGPT~\cite{yang2024omnimotiongpt} leverages prior human motion knowledge to generate animal motions, while Motion Avatar~\cite{zhang2024motionavatar} mitigates data scarcity via SinMDM-based augmentation and constructs Zoo-300K together with an LLM-agent framework for text-driven avatar and motion generation. However, such augmentation—built upon limited original datasets—cannot introduce motion patterns absent from the sources. Moreover, uniform augmentation may smooth out species-specific behavioral differences, limiting the ability to capture fine-grained motion dynamics.

AniMo~\cite{wang2025animo} introduces a Joint-Aware Spatiotemporal Encoder and Species-Aware Feature Modulation to accommodate morphological variations, along with the AniMo4D dataset. Yet this design implicitly assumes all species share the same joint configuration, simplifying learning but normalizing critical topological differences. As a result, AniMo can only generate motions for predefined species conditioned on text and cannot synthesize motions for arbitrary skeletons—a limitation for downstream flexibility. AnyTop~\cite{gat2025anytop} is the first to address motion generation across heterogeneous skeletal topologies, but its unconditional diffusion formulation prevents semantic control, restricting practical usability. Concurrent work~\cite{lee2025how} incorporates text conditioning into diffusion but still depends on the small Truebones Zoo~\cite{studios2022truebones} dataset, which severely limits generalization.

In summary, existing approaches either rely on fixed topologies with species-aware modeling or flexible topologies without controllability, and none achieve both large-scale and high-fidelity data. These gaps motivate our goal of developing a unified, text-driven motion generation framework capable of producing natural, semantically aligned motions across arbitrary animal skeletal topologies.

%% file: sec/3_dataset.tex
\section{Dataset}
\label{subsec:dataset}
We now introduce \textbf{OmniZoo}, a large-scale animal motion dataset featuring diverse species and rich multimodal annotations.
This dataset provides a comprehensive foundation for studying cross-species motion generation and serves as a valuable resource for developing generalized, topology-aware motion models.

\noindent\textbf{Data Collection: }We construct the dataset from two complementary sources to ensure comprehensive coverage of animal locomotion patterns. 
The first component is derived from the Truebones Zoo dataset~\cite{studios2022truebones}, which provides motion capture data spanning diverse animal categories, including mammals, birds, insects, dinosaurs, fish, and reptiles. 
The second component leverages the simulation game \textit{Planet Zoo}, which features high-fidelity animal models and motion datasets. 
We employ \texttt{cobra-tools} as the core extraction utility to obtain animal models and their corresponding animation sequences from the game engine.

\begin{figure*}[t!]
  \centering
  \includegraphics[width=\linewidth]{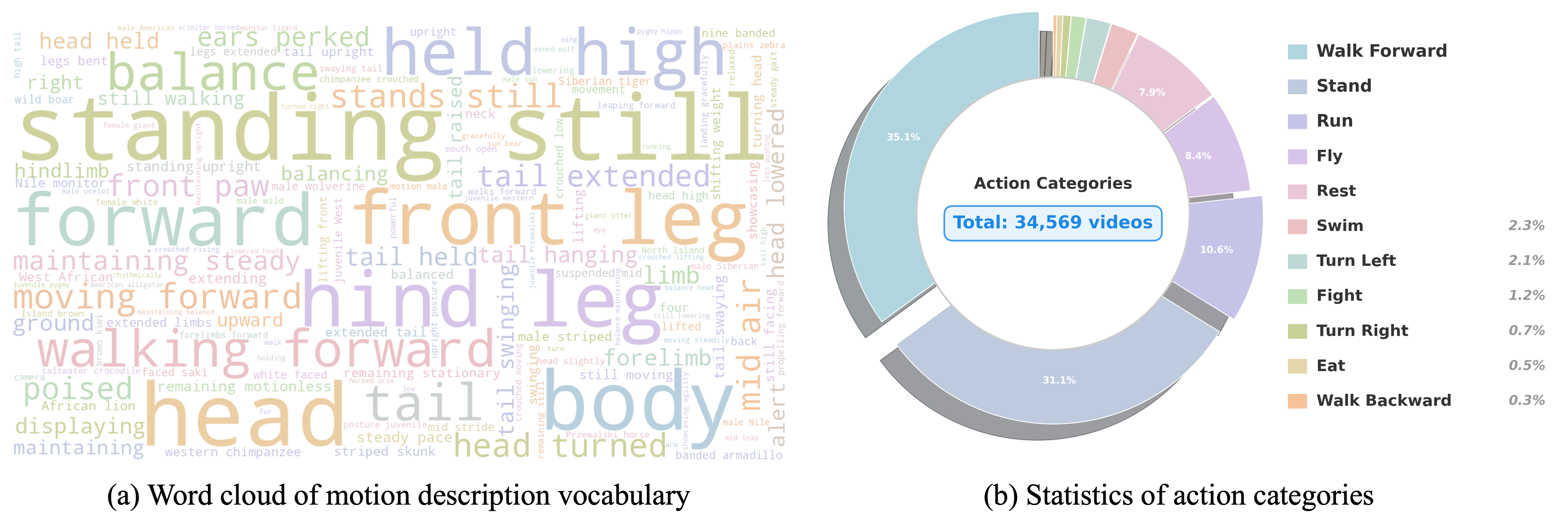}
    \caption{Overview of our motion dataset.}
  \label{fig:dataset}
\end{figure*}

\noindent\textbf{Data Statistics: }The OmniZoo dataset comprises \textbf{32,979 motion sequences} totaling \textbf{2,488,539 frames} across \textbf{140 species}. 
Motion sequence lengths range from $20$ to $240$ frames, with an average of $78$ frames per sequence. The skeletal structures exhibit significant heterogeneity, with an average of $133$ joints per skeleton.

We organize the dataset into four primary taxonomic categories based on skeletal topology and locomotion dynamics: \textit{aquatic animals} , \textit{quadrupeds} , \textit{birds}, and \textit{others}. 
The dataset is partitioned into training and test sets with a 0.95:0.05 split ratio. 
All motion data follows the $(T, d)$ format, where $T$ denotes sequence length and $d$ represents motion feature dimensionality, encompassing joint positions, rotations, and velocities. Table~\ref{tab:dataset_comparison} presents a comprehensive comparison with existing animal motion datasets.
Compared with these datasets, our dataset features heterogeneous skeletal motions from $140$ different species, along with rich multimodal annotations.



\noindent\textbf{Data Processing: }To ensure high-quality data, we implement a processing pipeline consisting of four key stages:

\textit{Mesh and Motion Data Extraction. }We extract high-quality 3D animal models directly from the original data sources, preserving the native topological configurations and joint connectivity for each species.

\textit{Motion Alignment. }We implement a comprehensive preprocessing pipeline to ensure data quality and consistency. 
All skeletons are normalized to a uniform scale, sequences are aligned to a consistent forward-facing direction, and root positions are centered at the origin. 

\textit{Video Generation. }Motion sequences are rendered in a game engine environment using the extracted 3D models, generating synchronized videos for each motion sequence.

\textit{Text Annotation. }We leverage state-of-the-art large language models, including Qwen~\cite{yang2025qwen3}, VideoLLaMA3~\cite{zhang2025videollama}, and LLaVA~\cite{lin2024video}, to perform comprehensive video understanding and motion analysis. 
Each motion sequence is processed through multiple LLMs to generate diverse descriptions that capture both \textbf{low-level kinematic features} and \textbf{high-level semantic information}. 
Fig.~\ref{fig:dataset} illustrates the rich range of motion descriptions in our dataset, spanning from concise summaries to fine-grained, detailed narratives.


\begin{table}[t]
\centering
\caption{Comparison with existing animal motion datasets.}
\label{tab:dataset_comparison}
\begin{threeparttable}  
\setlength{\tabcolsep}{3pt}
\small
\begin{tabular}{l@{\hspace{4pt}}c@{\hspace{4pt}}c@{\hspace{4pt}}c@{\hspace{4pt}}c@{\hspace{4pt}}c@{\hspace{4pt}}c}
\toprule
\textbf{Dataset} & \textbf{\#Sp.} & \textbf{\#Seq.} & \textbf{\#Fr.} & \textbf{\#Txt.} & \textbf{Topo.} & \textbf{Modal.} \\
\midrule
Truebones Zoo~\cite{studios2022truebones} & 70 & 1.2K & 147K & -- & Het. & M+V+Me \\
AniMo4D~\cite{wang2025animo} & 114 & 78K & -- & 185K & Uni. & T+M \\
Zoo-300K~\cite{zhang2024motionavatar} & 65 & -- & -- & 300K & Het. & T+M \\
\midrule
\textbf{OmniZoo} & \textbf{140} & \textbf{33K} & \textbf{2.5M} & \textbf{46K} & \textbf{Het.} & \textbf{T+M+V+Me} \\
\bottomrule
\end{tabular}
\begin{tablenotes}
\footnotesize
\item Sp.: Species; Seq.: Sequences; Fr.: Frames; Txt.: Texts; Topo.: Topology; Modal.: Modalities; Het.: Heterogeneous; Uni.: Unified; T: Text; M: Motion; V: Video; Me: Mesh.
\end{tablenotes}
\end{threeparttable}  
\end{table}

%% file: sec/4_method.tex
\section{Methodology}
\label{subsec:method}

Our objective is to enable controllable motion generation conditioned on both textual descriptions and skeletal structures. 
Specifically, given a text prompt and a skeleton of arbitrary topology, our model synthesizes realistic motions for the input skeleton that faithfully match the semantic intent of the text. 
To support generalization across heterogeneous skeletons, we propose a motion generation framework built upon masked modeling. 
Fig.~\ref{fig:pipeline} provides an overview of our framework.
The entire training pipeline is divided into two stages. 
In Sec.~\ref{sec:stage1}, we develop the generalized motion residual VQ-VAE to discretize motion sequences from arbitrary skeletal topologies into a unified tokenized latent space. 
In Sec.~\ref{sec:stage2}, we introduce the topology-aware skeleton embedding module, which our key technical contribution.
It encodes the skeleton input into compact topology-aware embeddings. 
The topology-aware embeddings and text embeddings are jointly projected into the motion latent space through a masked transformer and a residual transformer. 
During inference, both text and skeleton inputs are converted into motion tokens, which are decoded by the residual VQ-VAE to produce the final motion sequence.

\begin{figure*}[h!]
  \centering
  \includegraphics[width=\linewidth]{figs/pipeline.png}
  \caption{Overview of our pipeline. Our network operates in two main stages: In the first stage, we extend the residual VQ-VAE into a generalized formulation capable of handling motion sequences from arbitrary skeletal topologies. In the second stage, we introduce text prompts and target skeletons as joint conditioning signals. We design a topology-aware skeleton embedding module that extracts both spatial geometry and structural topology from any given skeleton. These features are then fused with text embeddings and fed into a two-stage Transformer to achieve generalized conditional motion generation across diverse species and skeletal structures.}
  \label{fig:pipeline}
\end{figure*}

\subsection{Generalized Motion Residual VQ-VAE}\label{sec:stage1}

To handle heterogeneous skeletal topologies with varying joint numbers and structures, we propose a Generalized Motion Residual VQ-VAE extended from~\cite{guo2024momask}, which enables motion representation learning across arbitrary skeletons through joint-level padding, masking, and residual quantization. 
Fig.~\ref{fig:pipeline}(a) shows the overview of our design.
Given a motion sequence represented as $\mathbf{X} = \{\mathbf{x}_t\}_{t=1}^{T}$, where each frame $\mathbf{x}_t \in \mathbb{R}^{J_t \times d}$ encodes $J_t$ joints with per-joint motion features. 
Since $J_t$ varies across skeletons, we pad each sequence to the maximum joint number $J_{\max}$ in the batch, resulting in $\tilde{\mathbf{X}} \in \mathbb{R}^{T \times J_{\max} \times d}$. 
A binary mask $\mathbf{M} \in \{0,1\}^{T \times J_{\max}}$ is constructed, where $\mathbf{M}_{t,j}=1$ indicates valid joints and $0$ denotes padding. 
The mask is propagated through all network layers and losses, ensuring that only valid joints contribute to learning, thereby maintaining anatomical consistency and preventing bias from variable topologies.

The encoder $E(\cdot)$ extracts the masked latent features $\mathbf{Z}_0 = E(\tilde{\mathbf{X}}, \mathbf{M}) \in \mathbb{R}^{T \times J_{\max} \times d_z}$. 
Instead of a single-step quantization, we employ a residual vector quantization~\cite{guo2024momask, wang2025animo} hierarchy that progressively refines latent representations. 
At each quantization level $l$, a learnable codebook $\mathcal{C}_l = \{\mathbf{c}_{l,k}\}_{k=1}^{K_l}$ is used to quantize the residuals via nearest-neighbor lookup:
\begin{align}
\mathbf{R}_{l+1} = \mathbf{R}_l - \hat{\mathbf{R}}_l, \quad 
\hat{\mathbf{R}}_l = Q_l(\mathbf{R}_l), \quad 
\mathbf{R}_1 = \mathbf{Z}_0,
\end{align}
where $Q_l(\cdot)$ denotes the quantization operation at level $l$. 
The final quantized latent is the accumulated sum $\hat{\mathbf{Z}} = \sum_{l=1}^{L} \hat{\mathbf{R}}_l$. 
The decoder $D(\cdot)$ reconstructs the motion as $\hat{\mathbf{X}} = D(\hat{\mathbf{Z}}, \mathbf{M})$, with all losses computed only over valid joints indicated by $\mathbf{M}$.

To jointly ensure masked reconstruction and residual consistency, the overall training objective is defined as:
\begin{align}
\mathcal{L}_{\text{rvq}} &=
\frac{\sum_{t,j} \mathbf{M}_{t,j}\, \|\tilde{\mathbf{X}}_{t,j} - \hat{\mathbf{X}}_{t,j}\|_1}
{\sum_{t,j} \mathbf{M}_{t,j} + \epsilon} \nonumber \\
&\quad + 
\beta \sum_{l=1}^{L}
\frac{\sum_{t,j} \mathbf{M}_{t,j}\, \|\mathbf{R}_l - \mathrm{sg}[\hat{\mathbf{R}}_l]\|_2^2}
{\sum_{t,j} \mathbf{M}_{t,j} + \epsilon},
\label{eq:rvq_loss}
\end{align}
where $\mathrm{sg}[\cdot]$ denotes the stop-gradient operator and $\beta$ controls the residual commitment weight. 
The first term enforces topology-masked motion reconstruction, while the second term ensures hierarchical residual refinement across quantization levels. 

After training, each motion sequence can be represented by $V{+}1$ discrete motion token sequences $\mathbf{M} = [\mathbf{m}_v^{1:n}]_{v=0}^{V}$, where each $\mathbf{m}_v^{1:n} \in \{1, \ldots, |\mathcal{C}_v|\}^n$ corresponds to the ordered codebook indices of quantized vectors $\mathbf{b}_v^{1:n}$, satisfying $\mathbf{b}_{v,i} = \mathcal{C}_v[\mathbf{m}_{v,i}]$ for $i \in [1,n]$. 
Through this formulation, the model learns to compress heterogeneous skeletal motions into a unified discrete latent space. 

\subsection{Generalized Conditional Motion Generation}\label{sec:stage2}

In this stage, we aim to map the input text prompt and skeleton to the motion tokens learned from the previous stage.
We propose a topology-aware skeleton embedding module that encodes skeletons of arbitrary topology into structured token embeddings.
Together with text embeddings extracted from the prompt, these tokens are fed into a two-stage transformer to predict the corresponding motion tokens.

\subsubsection{Topology-aware Skeleton Embedding Module}

\begin{figure}[h!]
  \centering
  \includegraphics[width=\linewidth]{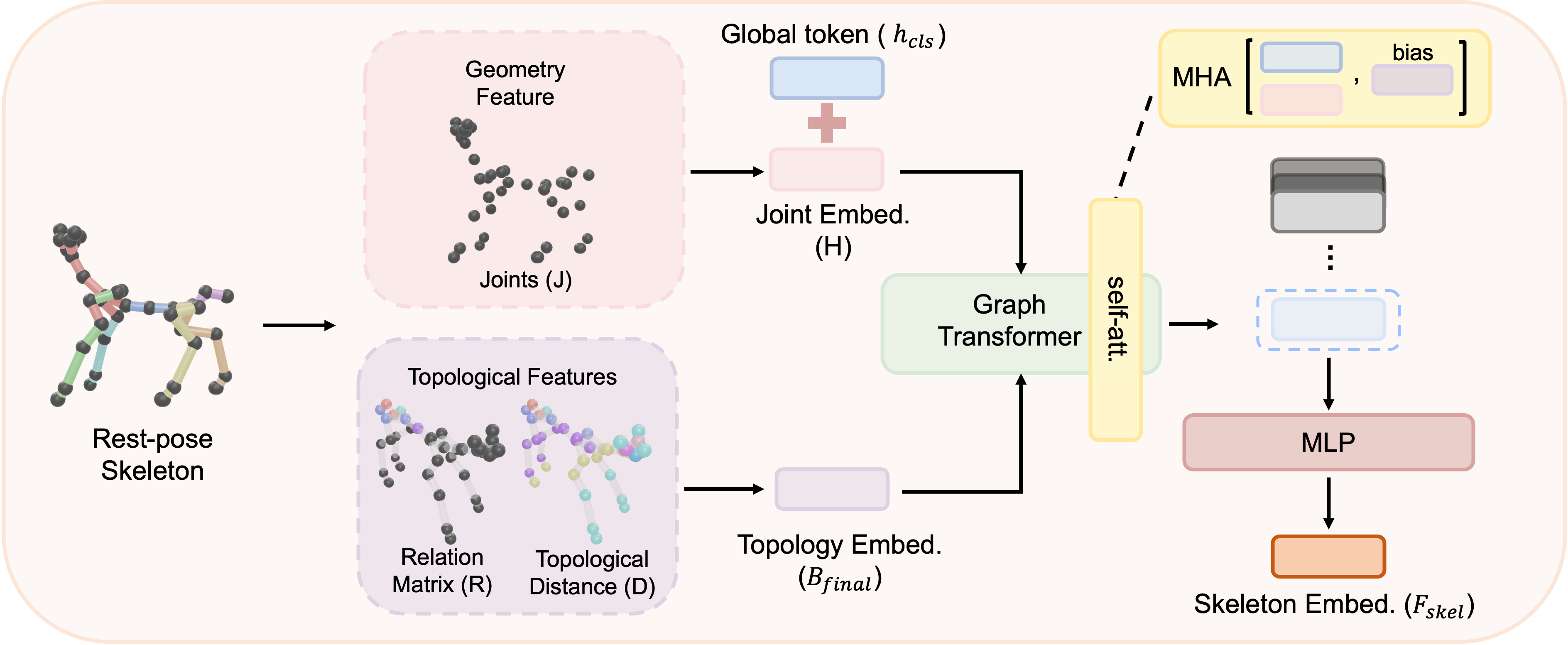}
  \caption{Overview of the topology-aware skeleton embedding module.}
  \label{fig:skeletonenc}
\end{figure}

To capture both geometric and structural cues from complex skeletons, we design a topology-aware skeleton embedding module that encodes the skeleton into a compact conditional embedding $\mathbf{f}_{\text{skel}} \in \mathbb{R}^{1 \times d_{\text{s}}}$. 
As illustrated in Fig.~\ref{fig:skeletonenc}, the encoder first embeds per-joint geometry, injects graph-based topology through attention biases, and then aggregates global semantic context via multi-head graph transformers. 

Given a skeleton with $K$ joints, we represent each sample with four components following~\cite{gat2025anytop}: 
the per-joint geometric feature $\mathbf{J} \in \mathbb{R}^{K \times d}$, 
the relation type matrix $\mathbf{R} \in \mathbb{R}^{K \times K}$ indicating semantic connectivity (self, parent, child, sibling, etc.), 
the shortest topological distance $\mathbf{D} \in \mathbb{R}^{K \times K}$ encoding graph structure, 
and a spatial mask $\mathbf{M} \in \{0,1\}^{K}$ specifying valid joints.
Each joint feature is projected into a latent feature space via a lightweight MLP:
\begin{equation}
\mathbf{H} = f_{\text{MLP}}(\mathbf{J}) = W_2\,\text{GELU}(W_1\mathbf{J} + b_1) + b_2,
\label{eq:mlp}
\end{equation}
where $W_1, W_2, b_1, b_2$ are learnable parameters.
A learnable global token $\mathbf{h}_{\text{cls}} \in \mathbb{R}^{1 \times d}$ is prepended to summarize global context, forming the input sequence:
\begin{equation}
\mathbf{Z}_0 = [\,\mathbf{h}_{\text{cls}},\, \mathbf{H}\,]^{\top} \in \mathbb{R}^{(K+1) \times d}.
\label{eq:concat}
\end{equation}

To inject topological priors into attention computation, we design a graph-aware attention bias based on topological distances and relation types. 
Specifically, $\mathbf{D}$ and $\mathbf{R}$ are mapped into head-dependent embeddings and summed with the extended spatial mask $\mathbf{M}'$, giving:
\begin{equation}
\mathbf{B}_{\text{final}} = \text{Emb}_{\text{dist}}(\mathbf{D}) + \text{Emb}_{\text{rel}}(\mathbf{R}) + \mathbf{M}'.
\label{eq:attnbias}
\end{equation}
We then feed $\mathbf{Z}_0$ into a stack of $L$ graph-transformer layers, where each layer consists of multi-head self-attention and a feed-forward network with residual connections:
\begin{equation}
\mathbf{Z}'_l = \text{MHA}(\mathbf{Z}_{l-1}, \mathbf{B}_{\text{final}}) + \mathbf{Z}_{l-1}, \quad
\mathbf{Z}_l = \text{FFN}(\mathbf{Z}'_l) + \mathbf{Z}'_l.
\label{eq:transformer}
\end{equation}
The multi-head attention incorporates the structural bias into the attention logits:
\begin{equation}
\text{MHA}(Q,K,V) = \text{softmax}\!\left(\frac{QK^{\top}}{\sqrt{d/H}} + \mathbf{B}_{\text{final}}\right)V,
\label{eq:mha}
\end{equation}
where $H$ denotes the number of attention heads. 
After $L$ layers, the contextualized CLS token is extracted and projected to the output dimension:
\begin{equation}
\mathbf{f}_{\text{skel}} = W_{\text{out}}\cdot \text{LayerNorm}(\mathbf{Z}_L[0]) + b_{\text{out}}, \quad 
\mathbf{f}_{\text{skel}} \in \mathbb{R}^{1 \times d_{\text{s}}}.
\label{eq:output}
\end{equation}
This conditional feature $\mathbf{f}_{\text{skel}}$ effectively encodes both semantic and topological priors of the skeleton, providing a rich, globally contextualized representation for controllable skeleton generation.

\subsubsection{Generalized Masked and Residual Transformer}

After obtaining the topology-aware skeleton embedding $\mathbf{f}_{\text{skel}}$ and the text embedding $\mathbf{f}_{\text{text}}$, we condition motion generation on both modalities. Concretely, a textual description $\mathcal{T}$ may consist of a species/category (e.g., \textit{cat}, \textit{human}), a motion class (e.g., \textit{walk}, \textit{jump}), and a fine-grained phrase (e.g., \textit{slowly turning the head while walking}). We first feed $\mathcal{T}$ into a SigLip2 encoder~\cite{tschannen2025siglip} to obtain a dense text feature $\mathbf{f}_{\text{text}} \in \mathbb{R}^{d_t}$. At the same time, the skeletal graph $\mathcal{S}=(\mathcal{V},\mathcal{E})$ is encoded by the above topology-aware skeleton embedding module to produce $\mathbf{f}_{\text{skel}} \in \mathbb{R}^{d_s}$. We then fuse the two modalities through a lightweight conditioning MLP:
\begin{equation}
\mathbf{f}_{\text{cond}} = \mathrm{MLP}\big([\mathbf{f}_{\text{text}} ; \mathbf{f}_{\text{skel}}]\big),
\end{equation}
where $[\cdot;\cdot]$ denotes concatenation. The resulting $\mathbf{f}_{\text{cond}}$ serves as a unified conditioning vector for both the masked transformer and the residual transformer, so that the model can leverage textual intent and skeletal topology simultaneously.

Built upon the discrete motion representations learned by our generalized motion residual VQ-VAE, we adopt a bidirectional masked transformer $G_{\theta}$ to model the base-layer motion tokens $\mathbf{m}^0_{1:n} \in \mathbb{R}^n$ (cf. Fig.~\ref{fig:pipeline}(b)). We randomly mask a variable fraction of tokens by replacing them with a special $[\text{MASK}]$ token, and let $\tilde{\mathbf{m}}^0$ denote the masked sequence. The training objective is to recover the masked tokens conditioned on $\mathbf{f}_{\text{cond}}$, i.e.,
\begin{equation}
\mathcal{L}_{\text{mask}} =
\sum_{\tilde{m}^0_k=[\text{MASK}]}
- \log p_\theta \left( m^0_k \,\middle|\, \tilde{\mathbf{m}}^0, \mathbf{f}_{\text{cond}} \right),
\label{eq:mask_cond}
\end{equation}
which is exactly the masked modeling loss in~\cite{guo2024momask} but augmented with our novel dual-condition input. 
Intuitively, the masked transformer learns to inpaint motion tokens in a way that is consistent with both the target skeleton and the described behavior.

To further model finer-grained motion details from the $V$ residual quantization layers, we additionally learn a residual transformer $R_{\phi}$ that shares the same backbone but owns $V$ separate embedding layers. 
During training, we randomly select a quantizer layer $j \in [1, V]$, embed all preceding layers $\mathbf{m}^{0:j-1}$, and sum them as the token input. Given the token embeddings, the conditioning vector $\mathbf{f}_{\text{cond}}$, and the layer indicator $j$, the residual transformer $p_\phi$ predicts the $j$-th layer tokens in parallel:
\begin{equation}
\mathcal{L}_{\text{res}} =
\sum_{j=1}^{V} \sum_{i=1}^{n}
- \log p_\phi \left( m^j_i \,\middle|\, m^{1:j-1}_i, \mathbf{f}_{\text{cond}}, j \right).
\label{eq:res_cond}
\end{equation}
We share the parameters between the $j$-th prediction head and the $(j{+}1)$-th motion token embedding layer to improve learning efficiency. 
Conditioned on $\mathbf{f}_{\text{cond}}$, this two-stage transformer stack learns to synthesize temporally coherent and semantically aligned motion token sequences that adapt to arbitrary skeletal topologies; the resulting discrete tokens are finally fed into the decoder of our generalized motion residual VQ-VAE to obtain the full motion in joint space.

\subsection{Inference}
During inference, given a text–skeleton condition $\mathbf{f}_{\text{cond}} = \mathrm{MLP}([\mathbf{f}_{\text{text}};\mathbf{f}_{\text{skel}}])$, we first initialize all base-layer tokens as $[\text{MASK}]$, i.e., $\mathbf{m}^0 = [\text{MASK}]$. 
The masked transformer $G_{\theta}$ predicts the base-layer motion tokens as $\hat{\mathbf{m}}^0 = G_{\theta}(\mathbf{m}^0, \mathbf{f}_{\text{cond}})$. 
For each subsequent residual quantization layer $j \in \{1, 2, \ldots, V\}$, the residual transformer $R_{\phi}$ predicts the $j$-th layer tokens based on the previously predicted layers and the same conditioning vector:
\begin{equation}
\hat{\mathbf{m}}^{j} = R_{\phi}(\hat{\mathbf{m}}^{1:j-1}, \mathbf{f}_{\text{cond}}, j).
\end{equation}
After obtaining all layers, the full quantized motion token sequence is $\hat{\mathbf{m}}^{1:V} = [\hat{\mathbf{m}}^0, \hat{\mathbf{m}}^1, \ldots, \hat{\mathbf{m}}^V]$. 
The corresponding quantized motion embedding $\hat{\mathbf{h}}_q$ is retrieved from the pre-trained codebooks, and the final motion sequence $\hat{\mathbf{X}}$ is reconstructed via the decoder of our generalized motion residual VQ-VAE.
This inference procedure enables controllable and topology-aware motion synthesis for arbitrary skeletons under text guidance.

%% file: sec/5_experiments.tex
\section{Experiments}

\begin{table*}[htbp]
\centering
\caption{Quantitative comparison of text-to-motion generation performance among different methods.}
\label{tab:comparison}
\small
\begin{tabular}{lcccccccc}
\toprule
\textbf{Method} & \textbf{FID↓} & \textbf{Diversity→} & \textbf{MatchingScore↓} & \textbf{MultiModality→} & \textbf{R@1↑} & \textbf{R@2↑} & \textbf{R@3↑} \\
\midrule
Real & - & 1.412 & 1.029 & - & 0.684 & 0.860 & 0.927 \\
MoMask~\cite{guo2024momask} & 0.085 & 1.392 & 1.181 & 0.513 & 0.454 & 0.625 & 0.712 \\
MMM~\cite{pinyoanuntapong2024mmm} & 0.084 & 1.395 & 1.141 & 0.032 & 0.503 & 0.680 & 0.766 \\
Animo~\cite{wang2025animo} & 0.141 & 1.377 & 1.226 & \textbf{0.724} & 0.337 & 0.473 & 0.546 \\
\rowcolor[HTML]{E7E6E6} 
Ours & \textbf{0.044} & \textbf{1.404} & \textbf{1.072} & 0.569 & \textbf{0.621} & \textbf{0.805} & \textbf{0.877} \\
\bottomrule
\end{tabular}
\end{table*}

\begin{figure*}[t!]
  \centering
  \includegraphics[width=\linewidth]{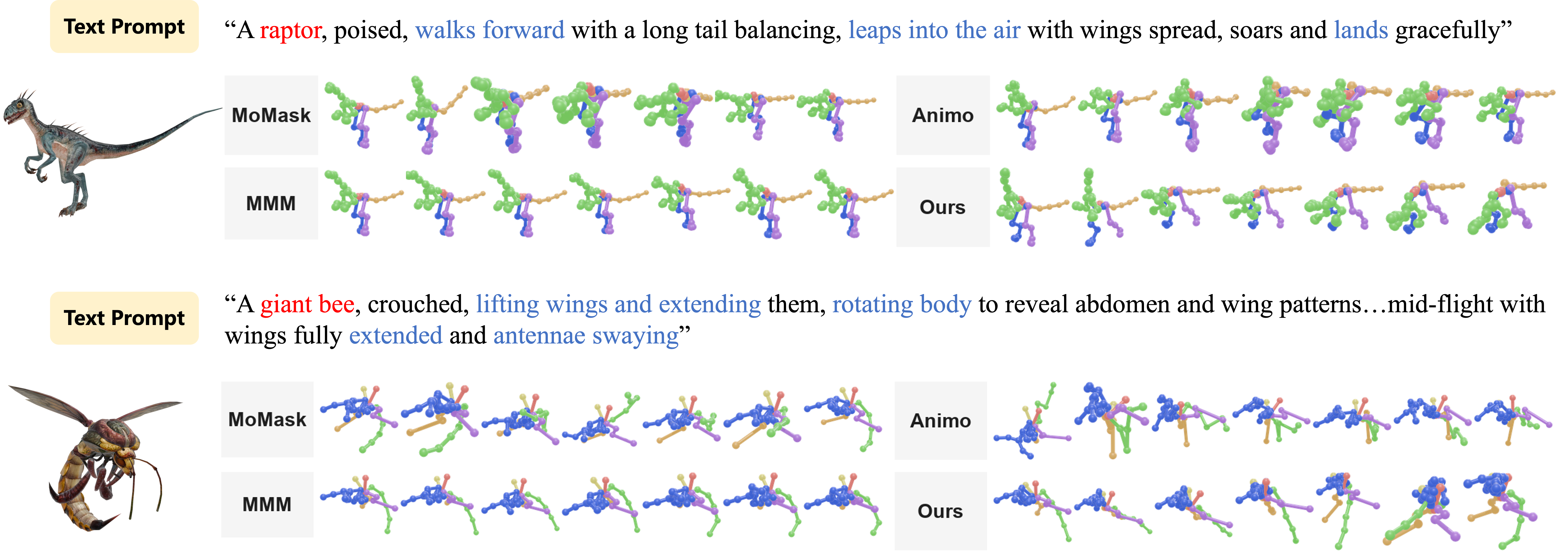}
    \caption{\textbf{Comparison of motion generation results on our testset.} Our model generates motions faithful to both the text prompt and the skeletal anatomy for heterogeneous skeletons, outperforming all baselines.
}
  \label{fig:baselines}
\end{figure*}

\subsection{Implementation Details}
Our generalized motion VQ-VAE employs residual vector quantization with $6$ layers, a $512$-dimensional codebook, and a temporal downsampling factor of $2$. Both generalized Masked and Residual Transformer use $8$ transformer layers with $4$ attention heads. 
Training uses AdamW optimizer with learning rate $1\times10^{-3}$ and the batch size is set to $256$. We train our model for $100$ epochs on motion sequences at $20$ FPS with a maximum $240$ frames. 
Classifier-free guidance uses a dropout probability of $0.1$ during training. 
All experiments are conducted on a single NVIDIA RTX $3090$ GPU with approximately $4$ days of training time per transformer model.

\subsection{Evaluation}
\noindent\textbf{Evaluation metrics.} We evaluate motion generation quality using seven metrics grouped into three categories. 
Quality metrics include \textbf{FID}, which measures the distributional distance between generated and real motions, reflecting overall realism. 
Semantic alignment metrics include \textbf{MatchingScore} (text–motion embedding similarity) and \textbf{R-Precision} (R@1/2/3), which assess how well generated motions correspond to textual descriptions. 
Diversity metrics include \textbf{Diversity} (inter-prompt variation) and \textbf{MultiModality} (intra-prompt variation), capturing the richness of motion outputs across and within prompts. While multimodality is desirable, it must be interpreted jointly with quality metrics to ensure that diversity does not come at the cost of semantic fidelity or motion realism.

\noindent\textbf{Baselines.} To the best of our knowledge, no existing publicly available model supports text-driven motion generation for \emph{heterogeneous} animal skeletons. We therefore compare against three representative baselines: MoMask~\cite{guo2024momask}, MMM~\cite{pinyoanuntapong2024mmm}, and AniMo~\cite{wang2025animo}.
Since MoMask and MMM assume a fixed SMPL template, we extend them to heterogeneous skeletons using joint padding and binary masking, and append species tags to text prompts for explicit conditioning. We also apply padding to meet AniMo's requirement for homogeneous skeletons and retrain it on OmniZoo following the official protocol.
AnyTop~\cite{gat2025anytop} is excluded because it performs unconditional generation without text control.
All baselines are trained with the same $95/5$ train–validation split and evaluated on the identical held-out motion sequences across all $140$ species in OmniZoo.

\subsection{Quantitative Results}

As shown in Tab.~\ref{tab:comparison}, our method achieves the best overall quantitative performance across all metrics.
First, the FID score measures the distributional similarity between generated motions and real motions. Our model attains the lowest FID, outperforming the best baseline MoMask by $48.2\%$, indicating that our generated motions are significantly closer to the ground-truth distribution in both quality and dynamics.
For semantic alignment, the MatchingScore and R@k metrics (R@1, R@2, R@3) evaluate how well the generated motion matches the input text. 
Our method achieves a $9.6\%$ lower MatchingScore and improves retrieval accuracy by $23.2\%$, $18.4\%$, and $14.5\%$ on R@1, R@2, and R@3, respectively, compared with the strongest baseline.
In terms of Diversity, our model reaches $1.404$, the closest to real data, demonstrating its ability to generate rich, varied motion patterns without mode collapse. 
The MultiModality score remains competitive—slightly lower than Animo—indicating that while our model balances diversity and coherence well, extreme stochasticity is intentionally controlled to maintain stronger semantic fidelity.

\subsection{Qualitative Results}

Fig.~\ref{fig:baselines} shows qualitative results where a target skeleton and a text prompt are jointly given. 
Baselines rely only on species labels and assume fixed, template-like skeletons during training; thus they fail to generalize to heterogeneous skeletal structures and often produce motions that do not match the prompt.

In contrast, our method explicitly encodes skeletal topology and geometry, enabling accurate motion synthesis even for unique species. 
For instance, in the giant bee example, the prompt "lifting wings and extending them, rotating body..." requires understanding both the fine-grained structure of the wings and the presence of the stinger. 
Only our approach successfully generates wing-lifting and extension behaviors while simultaneously producing the characteristic stinger rotation. 
These results align closely with the text description, whereas all baselines fail to capture these nuanced, species-specific motion semantics.
These qualitative trends are consistent with the quantitative results, confirming the superior semantic fidelity and structural awareness of our model.

\subsection{Ablation Study}
To validate the effectiveness of our model design, we conduct ablation studies on the core components, with results summarized in Tab.~\ref{tab:ablation}.
Quantitative metrics clearly show that removal of the topology-aware skeleton embedding module leads to a significant $54.6\%$ drop in FID and other alignment metrics, demonstrating that this module is essential for generating high-quality motions that remain semantically consistent with the input text across heterogeneous skeletal topologies.
Similarly, removing the motion-summary input also degrades $31.3\%$ FID performance, indicating that global motion context helps maintain temporal coherence and improves the fidelity of the generated sequences.

\begin{table}[htbp]
\centering
\caption{Ablation study on the testset.}
\label{tab:ablation}
\small
\begin{tabular}{lcccccccc}
\toprule
\textbf{Method} & \textbf{FID↓} & \textbf{R@1↑} & \textbf{R@2↑} & \textbf{R@3↑} \\
\midrule
wo. skeleton embed. & 0.097 & 0.402 & 0.558 & 0.634 \\
wo. motion summary & 0.064 & 0.617 & 0.789 & 0.800 \\
\rowcolor[HTML]{E7E6E6} 
Full Model & \textbf{0.044} & \textbf{0.621} & \textbf{0.805} & \textbf{0.877} \\
\bottomrule
\end{tabular}
\end{table}

\subsection{Application: Cross-Species Motion Transfer}

\begin{figure}[t!]
  \centering
  \includegraphics[width=\linewidth]{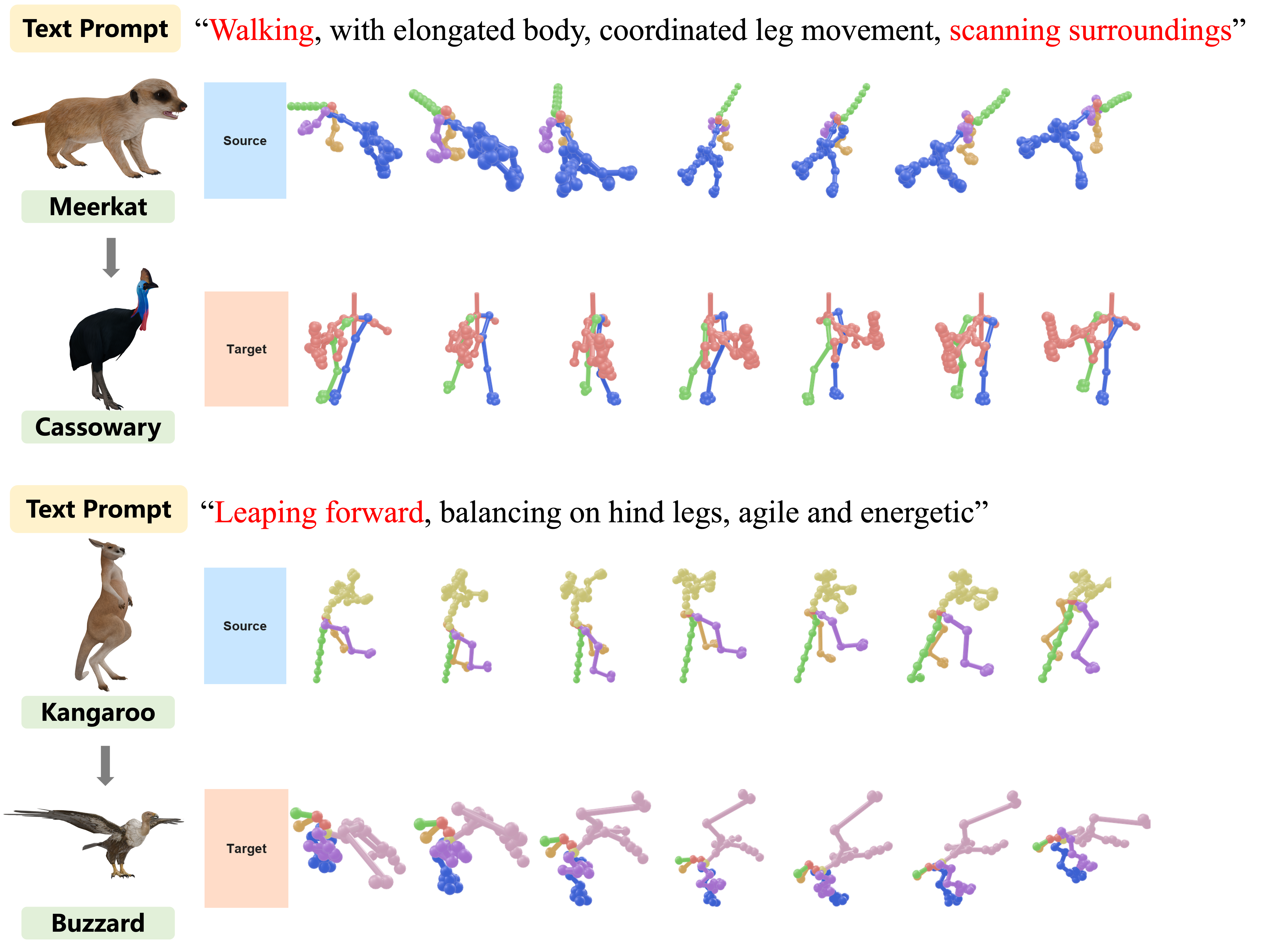}
    \caption{\textbf{Text-driven motion transfer for heterogeneous skeletons.} Our method generates biomechanically plausible motions for topologically distinct skeletons (e.g., meerkat and cassowary) from a single text prompt ("scanning surroundings").
}
  \label{fig:transfer}
\end{figure}

To further verify that our topology-aware skeleton embedding and multi-scale text conditioning enable the model to learn intrinsic correspondences between skeletal structure and motion semantics, we conduct a \textbf{cross-species motion transfer} experiment. 
Given the same text prompt, we apply it to skeletons of different species with distinct topologies. 
As illustrated in Fig.~\ref{fig:transfer}, when conditioned on the prompt "leaping forward", our model produces a kangaroo motion featuring clear leg compression–extension cycles and realistic airborne–landing transitions. 
When transferring the same prompt to a buzzard skeleton, the model not only preserves the global leaping behavior but also naturally incorporates wing flapping, reflecting the species-specific biomechanics. 
This experiment demonstrates strong out-of-distribution generalization and confirms that our method captures meaningful semantic–topology relationships beyond species-specific training patterns.

\subsection{Generalization to Real-World Animals}

\begin{figure}[t!]
  \centering
  \includegraphics[width=\linewidth]{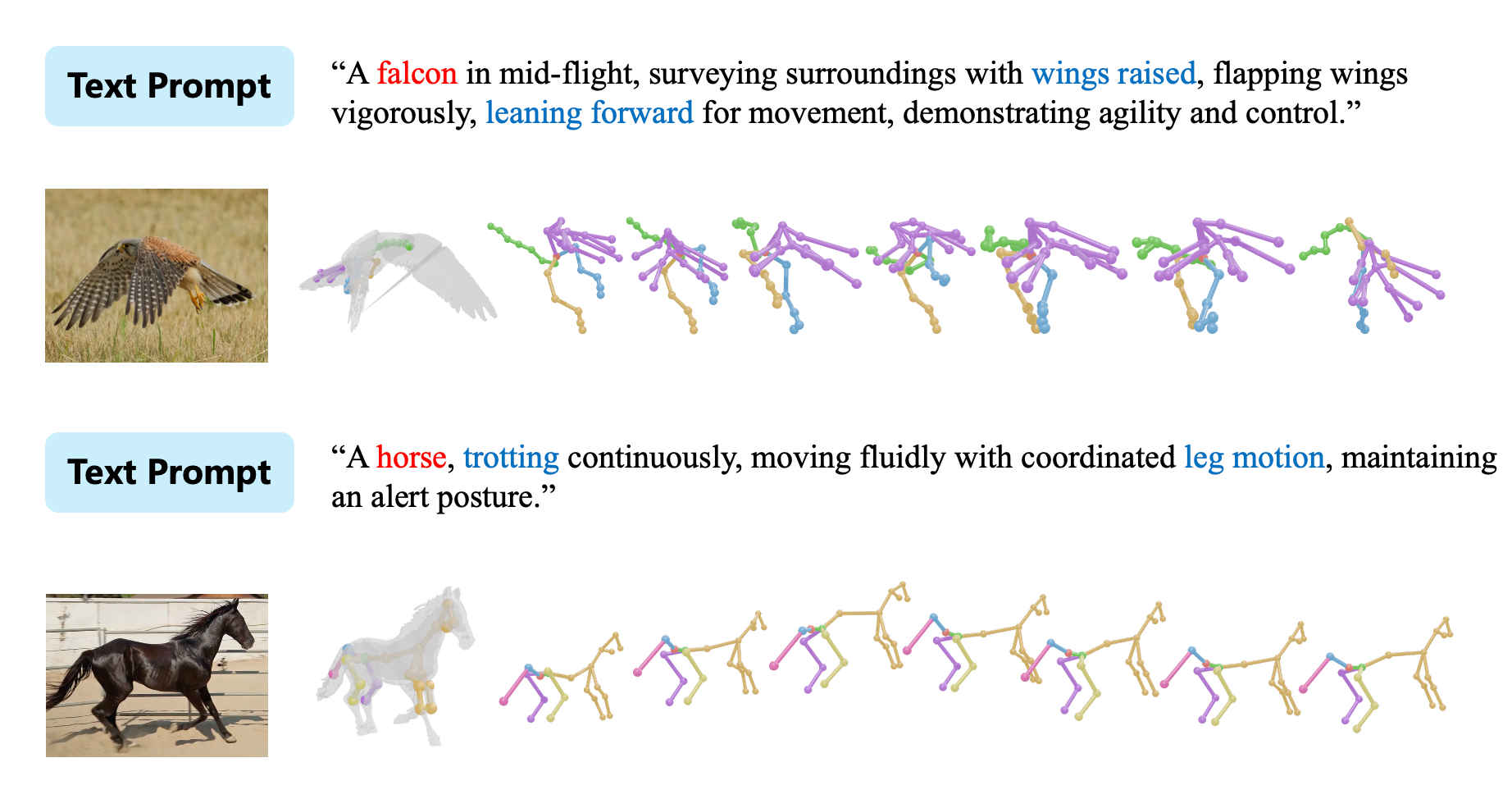}
  \caption{\textbf{Text-to-motion generation on real-world animals.} Our method successfully generates plausible motions for animals reconstructed from in-the-wild images. Top: A falcon performing aerial maneuvers with dynamic wing movements. Bottom: A horse executing a trotting gait with coordinated leg motion.}
  \label{fig:in-wild}
\end{figure}

To evaluate our method's practical applicability beyond curated datasets, we test on \textbf{real-world animal data} following a comprehensive reconstruction pipeline. 
We first collect diverse animal images from natural scenarios, then employ Hunyuan3D 2.0~\cite{hunyuan3d22025tencent} to reconstruct 3D meshes from 2D images. 
The generated meshes are rigged with skeletal structures using UniRig~\cite{zhang2025unirig}, producing articulated models. 
Finally, we apply our data normalization pipeline to convert the skeletons into a standardized representation compatible with our framework.

As shown in Fig.~\ref{fig:in-wild}, our model successfully generates realistic motion sequences for diverse real-world species. 
For the falcon example (top), given the text prompt \textit{"A falcon in mid-flight, surveying surroundings with wings raised, flapping wings vigorously, leaning forward for movement, demonstrating agility and control"}, our model produces natural aerial locomotion with appropriate wing dynamics and body posture. 
For the horse example (bottom), the prompt \textit{"A horse, trotting continuously, moving fluidly with coordinated leg motion, maintaining an alert posture"} yields a characteristic trotting gait with proper limb coordination and weight distribution.

These results demonstrate that our topology-aware approach generalizes effectively to real-world animal geometries beyond the training distribution, highlighting its potential for practical applications in wildlife animation, ecological simulation, and virtual character creation.

%% file: sec/6_conclusion.tex
\section{Conclusion, Limitation, and Future Work}\label{sec:conclusion}

In this paper, we present the first unified framework for text-driven motion generation across arbitrary skeletal topologies.
We introduce OmniZoo, a large-scale heterogeneous animal motion dataset covering $140$ species with rich multimodal annotations.
Built upon this foundation, our generalized autoregressive model—powered by a topology-aware skeleton embedding module—effectively fuses structural and semantic cues, enabling the synthesis of temporally coherent, physically plausible, and semantically aligned motions for diverse species.

Despite these advancements, our approach still has several limitations.
Currently, the model focuses on animal motion and cannot yet handle articulated human bodies or other complex character types.
In addition, our control signals are restricted to text and skeleton inputs, while richer modalities—such as video-based motion guidance—remain unexplored. 
Pursuing production-quality motion generation is another promising direction. 
We leave these challenges for future work.